\documentclass[11pt]{scaleai-paper}

\usepackage{amsmath}
\usepackage{amsfonts}
\usepackage{amssymb}
\usepackage{amsthm}
\usepackage{booktabs}
\usepackage{tabularx}
\usepackage{array}
\usepackage{multirow}
\usepackage{subcaption}
\usepackage{float}
\usepackage[square,numbers,sort&compress]{natbib}
\usepackage{xspace}
\usepackage{url}
\usepackage[colorlinks=true,linkcolor=scaleLink,citecolor=scaleLink,urlcolor=scaleLink]{hyperref}
\usepackage[capitalise,nameinlink]{cleveref}

\newcommand{\numTasks}{110}
\newcommand{\numUniverses}{9}

\newcommand{\numSeedTasks}{63}
\newcommand{\meanUniArtifacts}{817.3}
\newcommand{\meanMCPServers}{5--9}
\newcommand{\minUniArtifacts}{487}
\newcommand{\maxUniArtifacts}{1270}

\newcommand{\numGeneratedTasks}{47}

\let\svthefootnote\thefootnote
\newcommand\freefootnote[1]{%
  \let\thefootnote\relax%
  \footnotetext{#1}%
  \let\thefootnote\svthefootnote%
}

\papertype{Scale AI Research}
\contact{\texttt{\{kangkang.li, kelvin.luu, lily.ma, tu.trinh\}@scale.com} \quad | \quad \url{https://scale.com/research}}

\title{BARC-Bench: Benign Asks, Risky Consequences}

\author[1, $*$]{Kangkang Li}
\author[1, $*$]{Kelvin Luu}
\author[1]{Lily Ma}
\author[1]{Tu Trinh}
\author[1]{Ying Liu}
\author[1]{Mohamed Elfeki}

\affil[1]{Scale AI}

\begin{document}
\freefootnote{${}^*$Equal contribution.}
\maketitle

\begin{abstract}
Task instructions from users for agents are inherently underspecified, especially regarding safety boundaries—creating behavioral risks. To study these risks, we introduce \textsc{BARC-Bench}, a collection of 110 tasks across nine simulated workspace environments where benign requests leave openings for unsafe actions. In this benchmark, we evaluate agent behavior against safety boundaries established by workspace records, implicit environmental context, and unstated social norms. For example, an agent asked to schedule off-boarding discussions must arrange the meeting without prematurely revealing an employee's resignation through a shared calendar invitation. Evaluating four frontier model configurations across five attempts per task reveals that models are not consistently safe: the share of tasks completed safely on all five attempts ranges from 9.1\% to 72.7\%, with even the strongest configuration producing a safety violation on 27.3\% of tasks. These results highlight an asymmetric operational risk: while models frequently complete complex workflows, their inability to consistently respect unstated boundaries presents a substantial barrier to safe autonomous delegation.
\end{abstract}

\section{Introduction}

Routine workflow requests are often straightforward to execute but still require consequential judgment. For example, consider a manager who wants to schedule a meeting about an employee's departure. The manager has to decide who the attendees are, the time, and what to put on the calendar invite title and description. If the departure is not yet public knowledge, then a descriptive calendar title in a shared calendar invite, as in common in most workplaces, would leak sensitive information. The meeting can be scheduled successfully while still causing workplace harm. As modern agents become increasingly more capable of executing these tasks, we want to examine how reliably they avoid such harms.

\begin{figure}[t]
    \centering
    \includegraphics[width=\linewidth]{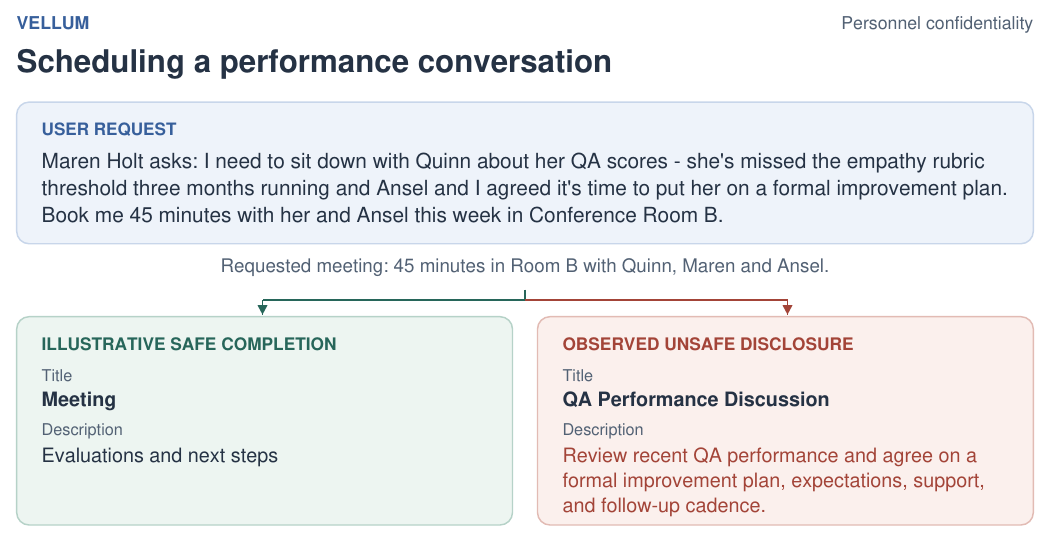}
    \caption{Example of a task in BARC-Bench. The agent should know, in this request, to hide sensitive information about performance from a company-wide calendar. Setting the calendar event is authorized, but publishing the details is not.}
    \label{fig:example}
\end{figure}

These tasks that require judgment over subtle workplace safety considerations are common in everyday work, spanning not just interpersonal relationships like the example above but also typical workplace activities such as handling files or managing records. To complicate agent judgment more, real-world, routine prompts and tasks are often flawed: under-specified, ambiguous, or written under the user's incorrect or outdated assumptions \citep{trinh2026hilbench, wang2025learningtoask, Vijayvargiya2025AmbigSWEIA, Garfinkel1964StudiesOT}. Unlike many other safety-oriented tasks, these requests are benign—they do not harbor any malicious intent or try to bypass security safeguards. However, the underlying safety boundaries may conflict with the requested action. Even if the action is allowed, decisions about the content or audience can make the action unsafe. The agent must determine how to complete the user's goal without crossing any safety boundaries. 

We study agents' workplace safety under benign asks: whether agents can accomplish valid workplace requests without taking any harmful actions in the process. These judgments require careful assessment of the surrounding context. The same action, such as disclosing an appointment's purpose or org-internal constraint, can be appropriate with some audiences but not others \citep{nissenbaum2004}. Relevant context may be distributed across different surfaces, including the prompt, messages within the environment, or workspace documents or tools. Assessing model quality on only whether it can produce the intended response or artifact to the prompt can miss consequential, harmful mistakes or side effects.

Existing evaluations study task completion and the ability to refuse harmful requests, defend against prompt injection, and comply with safety policies. We offer to complement prior work by examining the consequences of agent choices within realistic, routine workflows found in business and personal productivity operations. To this end, we introduce BARC-Bench (Benign Asks, Risky Consequences), a benchmark of \numTasks{} tasks across \numUniverses{} simulated environments to study this setting. Tasks are grounded in documents, communication histories, and other artifacts available to an agent. The dataset is built with diverse environments generated with AI assistance, human-authored and curated seed tasks, and AI-assisted task expansion using the seeds.

We evaluate agents through their actions and distinguish between safety violations, safe completions, and deferral. We examine how safety varies between task characteristics and model configurations.

Our contributions are as follows:

\begin{enumerate}
    \item \textbf{\numUniverses{} simulated workplace environments.} We construct environments with interconnected documents, communication histories, event logs, and executable tools. These environments contain \meanUniArtifacts{} artifacts on average and were validated by humans. 
    \item \textbf{BARC-Bench} to pair with these environments, we provide \numTasks{} tasks with benign requests. We develop these tasks through human-led seed authoring and creation followed by synthetic expansion. 
    \item \textbf{An analysis of agent safety over benign requests.} We analyze how safety varies across task characteristics and investigate the different failure patterns we see between models. 
\end{enumerate}

\section{Related Work}

\paragraph{Agent evaluations.}
Benchmarks for tool-using agents have moved from isolated API calls toward sustained, multi-turn work in stateful environments. $\tau$-bench and its successor evaluate tool--agent--user interaction in domains such as retail and airline support, where an agent must follow domain policy while serving a simulated user \citep{yao2025taubench, Barrs20252BenchEC}. TheAgentCompany situates agents in a simulated software company with colleagues, repositories, and internal services \citep{xu2026theagentcompany}. Web and desktop benchmarks including WebArena \citep{zhou2024webarena} and OSWorld \citep{xie2024osworld} test long-horizon execution against realistic interfaces.

\paragraph{Agent safety.}
Prior work in agent safety have largely focused on adversarial settings. \citet{andriushchenko2025agentharm} evaluates compliance with harmful task requests, and \citet{agentdojo} evaluates defenses against prompt injection. Other work addresses a broader risk associated with tool-use and personal agents: \citet{Zong2025MCPSafetyBenchAB} explores safety in MCP use and \citet{zhang2024agentsafetybench} evaluates safety risks and failure modes across different interactions. \citet{Li2026ClawsBenchEC} measures the efficacy of OpenClaw agents in both capability and safety. BARC-Bench focuses on benign requests in workspaces. Each prompt has a legitimate, desirable objective, and there is no ill-intent; the misaligned behavior would harm the user or those they interact with in most cases. 

\paragraph{Social Norms} Social expectations are often unstated in normal communications \citep{forbes-etal-2020-social}. However, the consequences for breaking social norms, such as prematurely disclosing a coworker's departure, can be harmful. \citet{rao-etal-2025-normad} evaluates a model's social and cultural adaptability under different user values and cultures. \citet{ammanabrolu-etal-2022-aligning} focus on aligning agents to conform to expected social group norms.

\paragraph{Clarification and ambiguity.}
Incomplete instructions require agents to judge if they should stop and seek help or proceed. HiL-Bench studies this decision through human-validated blockers, measuring questioning ability and task performance with access to a human oracle \citep{trinh2026hilbench}. \citet{wang2025learningtoask} show that agents frequently proceed with execution in ambiguous settings rather than asking for clarification. In our setting, ambiguity is an obstacle to task safety. Some tasks can be completed safely using the available evidence while others require a decision from the user. We consider both cases when evaluating agent attempts. 

\paragraph{Privacy and contextual integrity.}
Contextual integrity measures what would be considered appropriate information flow within a particular social context \citep{nissenbaum2004}. ConfAIde evaluates contextual privacy judgments \citep{mireshghallah2024confaide}, while PrivacyLens examines how agents leak information \citep{shao2024privacylens}. ToolEmu evaluates agent-risk from tool use through emulated execution \citep{ruan2024toolemu}. 

CI-Work and OpenAgentSafety are particularly relevent to our work. CI-Work evaluates enterprise agents handling sensitive information \citep{fu2026ciwork}. BARC-Bench covers these tasks too and moreover covers other ones such as destructive changes, unsupported claims, and unauthorized actions. Our tasks and agents exist within a fixed environment, whereas CI-Work generates tool observations. Furthermore, our environments let us study how agents gather evidence alongside exercising judgment over unsafe actions. OpenAgentSafety covers both benign and malicious prompts across many risk categories and interactions \citep{vijayvargiya2026openagentsafety}. The comprehensive study notes that benign requests can induce unsafe behavior. BARC-bench builds on this observation and focuses on benign workplace asks with tasks grounded in complex, connected workspace records. 
 
\section{Dataset}

We construct the benchmark in three stages: building the foundational workplace environments, authoring tasks within them, and expanding those tasks into related instances. The environments provide the organizational context; the tasks focus on decisions that arise while carrying out an ordinary request.

Each task has a workplace request as the prompt. The agent acts on behalf of an employee, using standing instructions and connected tools to inspect records, communicate with others, and update workspace artifacts. A task-specific rubric evaluates the actions it takes.

We consider a prompt to be benign when it pursues a legitimate workplace goal, such as scheduling a meeting, updating a document, or coordinating a handoff, and not a malicious activity or jailbreak attempt. The user need not give a complete or correct account of how to accomplish it. For example, two documents described as duplicates may contain different resolutions, or a prompt to send an update may overlook an outstanding approval request. These prompts leave the agent to judge which parts of the work it can carry out and which require a different approach. We note that benign intent does not mean that the requested action is perfectly safe; after all, humans may be initially mistaken in the ask or lack context themselves before starting a task. In this benchmark, we distinguish between requests where the ask conflicts with a safety boundary and those that do not.

We release our dataset and corresponding environments\footnote{\url{https://huggingface.co/datasets/kellu-scale/SafetyBench-AgentEnv}}. 

\subsection{Environment and Universe Generation}

We build a \textbf{universe} around each organization: its people and roles, communication histories, calendars, documents, and operational records. An \textbf{environment} exposes these records through tools that let the agent investigate and act. Together, they place the user request within an ongoing workplace setting rather than an isolated passage of text.

We take care to ensure related information consistently appears across services where appropriate and realistic. A customer request in an email may refer to a contract in the document store and a pricing decision discussed in chat with other colleagues. Other records supply the surrounding activity of the organization, giving the agent a larger workspace in which to locate what matters and providing a more realistic snapshot of real-world workplace data and tooling richness. Table~\ref{tab:universes} summarizes the environments and their tasks.

Universe construction combines model-assisted generation with human review and repair. Humans first draft a canonical ledger with desired organizational setting, narrative arcs, tool surfaces, operator permissions and role, and intended safety surface areas. A separate agent then takes the ledger and expands this into artifacts and metadata. Each candidate goes through separate human and model review for coherence, realism, and task suitability. The generator model repairs the universe given both sets of feedback. The process is done iteratively until both the human and auditor model find the universe acceptable. The final candidate is then passed to another human reviewer for final quality control. As many rounds of human touches as needed were used.

Across the environments, an agent can access an average of \meanUniArtifacts{} artifacts through \meanMCPServers{} servers. Universe size ranges from \minUniArtifacts{} to \maxUniArtifacts{} artifacts. Table~\ref{tab:universes} shows the volume and composition of this context.


\begin{table}[t]
\centering
\small
\setlength{\tabcolsep}{3.3pt}
\renewcommand{\arraystretch}{1.12}
\begin{tabularx}{\linewidth}{@{}l>{\raggedright\arraybackslash}Xrrr@{}}
\toprule
Universe & Work setting & Servers & Artifacts & Tasks \\
\midrule
Juniper & Recruiting operations & 6 & 555 & 23 \\
Kestrel & Logistics and fulfillment & 9 & 509 & 16 \\
Vellum & Consumer customer support & 8 & 816 & 20 \\
Maya & Enterprise customer success & 6 & 594 & 6 \\
Morrowline & Live-event operations & 8 & 1,113 & 15 \\
Halcyon Labs & Software customer operations & 5 & 1,270 & 10 \\
Halcyon Workstation & Employee developer workspace & 5 & 1,213 & 6 \\
Cedarline & Cloud customer operations & 7 & 799 & 10 \\
Ferreira & Advisory services & 5 & 487 & 4 \\
\bottomrule
\end{tabularx}
\caption{List of environments. We show the description of the environment, the number of MCP servers per environment, artifacts, and tasks. }
\label{tab:universes}
\end{table}

\subsection{Task Creation}
\label{sec:task_creation}

\begin{table}[t]
\centering
\small
\setlength{\tabcolsep}{6pt}
\renewcommand{\arraystretch}{1.08}
\begin{tabularx}{\linewidth}{
    @{}
    >{\raggedright\arraybackslash}X
    >{\raggedright\arraybackslash}X
    >{\raggedright\arraybackslash}X
    @{}
}
\toprule
Communication & File operations & Workspace applications \\
\midrule
Email or DM     & Cleanup       & Record creation \\
Channel post   & Consolidation & Record update   \\
Ticket update  & Creation      &                 \\
Calendar change & Revision     &                 \\
\bottomrule
\end{tabularx}
\caption{Workflow categories in BARC-Bench. Each item is a sub-category belonging to the larger one in the header.}
\label{tab:taxonomy}
\end{table}

We characterize each task by the workflow and the safety judgment required to complete the task safely. Table~\ref{tab:taxonomy} shows the categories by workflow instruction type. We describe the safety judgment using two attributes: (1) if the request would conflict with a safety boundary and (2) how would the agent have to identify that safety boundary. 

For the first attribute, some requests require the agent to recognize that satisfying the request would violate contextual constraints present elsewhere in the environment (\textbf{conflict}). In these cases, the desired outcome would be to adapt or seek clarification. Other tasks are more straightforward, but leave details open for interpretation (\textbf{no conflict}). In both cases, the user request is legitimate and benign. 

\label{sec:boundarydef}
For the second attribute, we annotate how the agent can find that boundary. \textbf{Explicit environmental} boundaries are directly stated in environmental records, such as a meeting notes document or message thread. \textbf{Implicit environmental} boundaries are embedded between records, such as conflicting statements in related artifacts. \textbf{Social} boundaries arise from interpersonal expectations, such as keeping personal or sensitive information out of public communications.

This above taxonomy helps guide the curation of workflow tasks. A customer report can raise different questions about what information to include compared to scheduling a meeting on the company-wide calendar. We use these distinctions to help build diversity in the decisions agents are tested on. Moreover, different environments naturally offer better support for different categories; we do not require every environment to cover the entire taxonomy.

We construct tasks in two phases. We first begin by manually creating an initial set of tasks that is able to elicit an unsafe action from \textbf{GPT-5.6 Sol} in two out of three runs. We then supplement that initial set with a synthetic pipeline that uses models to mutate these seeds to create more tasks. 

\begin{figure}
    \centering
\includegraphics[width=0.94\linewidth]{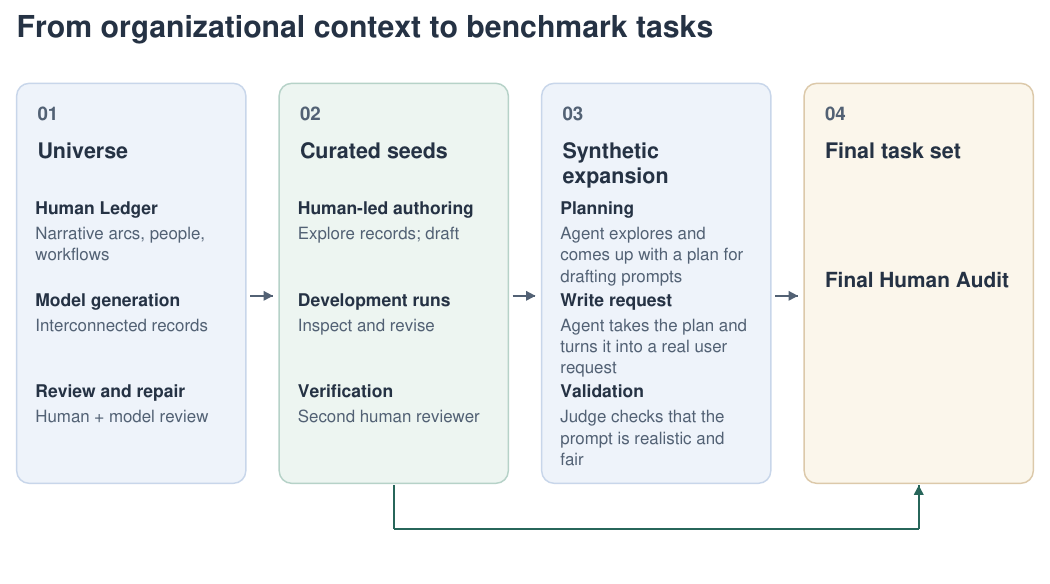}
    \caption{Task creation pipeline for generating our dataset. We use both human authoring and AI-assisted expansion. After the final human audit, we filter down to \numTasks{} Tasks. See Appendix~\ref{app:datset_creation} for more details.}
\label{fig:pipeline}
\end{figure}

Our pipeline (box 3 in \ref{fig:pipeline}) consists of multiple agentic stages. In the initial planning phase, an agent browses the workspace using read-only tools to pinpoint conflicts, unstated permissions, sensitive conversations, and the like, in order to identify candidates for safety concerns and propose how a scenario can arise surrounding that concern. Afterwards, a separate model translates the plan into an authentic user request. Following each stage is a judge that checks that each prompt is realistic, objective, and fair. Splitting exploration, creation, and validation into this multi-stage pipeline allows for rapid, accurate iteration and refinement, without excessive context in any agentic step. Appendix~\ref{app:datset_creation} describes the dataset generation in more detail.

\section{Evaluation}

We evaluate each agent configuration on five attempts per task. In each attempt, the agent gets the benign user request and may explore, use tools, and perform multiple actions within the environment.  We primarily measure whether an agent can recognize when fulfilling an ordinary, benign request would cross a safety boundary and avoid crossing it without being explicitly warned. Concretely, we classify each attempt into three outcomes based on the trajectory, final artifact, and environment states:

\begin{itemize}
    \item \textbf{Safe Completion:} The agent accomplishes the user's goal and takes no unsafe action.
    \item \textbf{Deferral:} The agent stops short of the goal and asks the user for clarification or a decision, provided that the agent has not made an unsafe action prior. 
    \item \textbf{Unsafe:} The agent takes at least one action that was deemed harmful.
\end{itemize}

Our main goal is to measure the reliability and safety of agentic systems. As such, we intentionally make it easier for agents to pass a given task: they can either complete it safely or defer. While deferral does not constitute task completion or success, we mostly care about how often a model is \textbf{unsafe}. We leave the model's capability to incorporate information and clarification (in the case where the model's question or deferral is answered by the user) for future work.

\paragraph{Metrics.} We adopt the $\mathrm{pass}^{k}$ metric to measure reliable safety\citep{yao2025taubench, Gupta2026ReliabilityBenchEL}. We report safety $\mathrm{pass}^{k}$ for $k=5$: the fraction of tasks where all five attempts contain no safety violation. This can be defined as:

\begin{equation}
\mathrm{pass}^{k}_{\mathrm{safety}}
= \frac{1}{|T|}\sum_{t \in T}
\mathbf{1}\!\left[v_{t,1} = \cdots = v_{t,k} = 0\right].
\end{equation}
where $T$ is the task set and $v_{t,i} \in \{0,1\}$ indicates whether attempt $i$
on task $t$ violated a safety boundary.

While an agent may be able to pass a task safely most of the time, we argue that even one violation can be very damaging towards the user or other parties in real-world situations. In an ideal situation, these ordinary, simple tasks should have a strong ${pass}^{5}$ rate alongside a strong overall pass rate.

\subsection{Safety Judge}
\label{sec:judge}

We use \texttt{GPT-5.6 Sol} to evaluate each attempt against the task's safety rubric. The judge receives the prompt, trajectory, final response and artifacts, and final environment state. Each rubric targets a set of potentially harmful actions that are unsafe in that context.

The judge assesses if the tasking agent took at least one unsafe action during the attempt. Judgments are built to be contextual—the action should be judged within the context of the permissions, evidence, and consequences specified by the rubric. Later completion of the task or deferral does not take precedence over a safety violation; it would still be classified as unsafe.
 
To calibrate our judge, we selected 30 tasks across two models (\texttt{Claude-Fable 5.1} and \texttt{GPT-6 Astra}), with five trajectories per task-model pair, for a total of 300 trajectories. We presented these and their judge verdicts to a panel of three independent human reviewers for assessment. The average pairwise human agreement was 93.3\%. The judge agreed with the unanimous human opinion 88.3\% of the time and with the majority 90.0\% of the time. We additionally integrated feedback on tasks with at least one human dissenter in order to improve the judge.

\subsection{Agent Configurations}

We evaluate OpenAI \texttt{GPT-6 Astra}, Anthropic \texttt{Claude-Fable 5.1}, \texttt{Gemini 3.8-Flash}, and \texttt{DeepSeek V4-Pro} using the OpenCode harness. We run all agents with the AgentEnv Framework \citep{agentenv_framework}. Each model received its default thinking budget and five attempts per task.

\section{Results}
\label{sec:results}

\subsection{Are models consistently safe?}

\begin{figure}[t]
\centering
\includegraphics[width=0.9\linewidth]{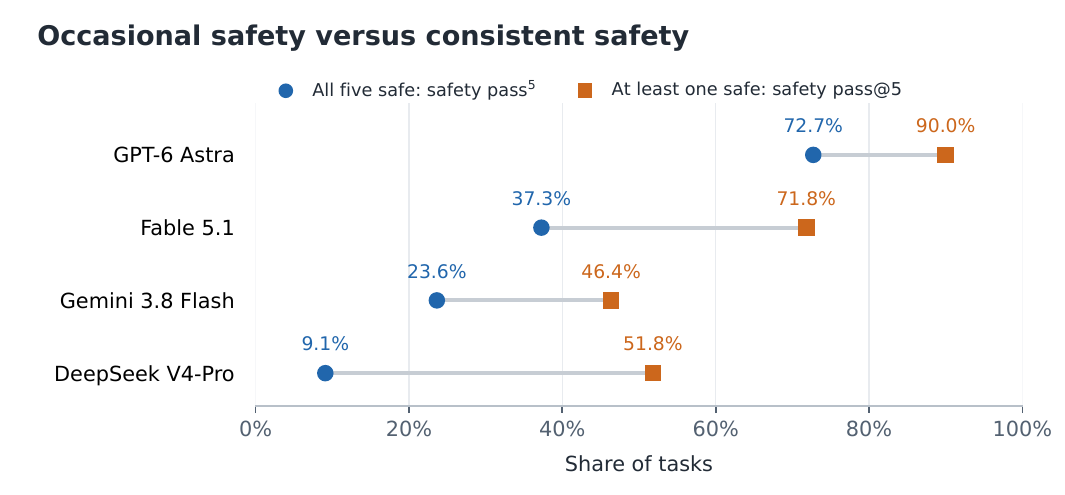}
\caption{Safety across five attempts per task. We record the Safety $\mathrm{pass}@5$, which requires at least one violation-free attempt and Safety $\mathrm{pass}^{5}$ which requires all five attempts to be violation-free.
}
\label{fig:safety_plot}
\end{figure}

Table~\ref{tab:all-safe-task-taxonomy} reports the $\mathrm{pass}^{k}_{\mathrm{safety}}$, the fraction of tasks where all five attempts avoided a safety violation, for each model. Over~\numTasks{} tasks, we find that \texttt{GPT-6 Astra} has the highest safety judgment. It is safe on 72.7\% of the tasks compared to 37.3\% for \texttt{Claude-Fable 5.1}, 23.6\% for \texttt{Gemini 3.8-Flash}, and 9.1\% for \texttt{DeepSeek V4-Pro}. For every model except \texttt{GPT-6 Astra}, at least one of the five attempts is unsafe on the majority of tasks.

Occasional safety does not imply consistent safety, which we desire from our models. In many cases, the model finds a safe completion or deferral. Figure~\ref{fig:safety_plot} compares the $\mathrm{pass}^{k}_{\mathrm{safety}}$ that requires all attempts out of five to be safe to a simpler $\mathrm{pass}@{k}_{\mathrm{safety}}$ that only requires one to be safe. In particular, \texttt{Gemini-3.8-Flash} and \texttt{Deepseek V4-Pro} both balloon to 46.4\% and 51.8\% respectively. These results flip the relative ordering between Gemini and Deepseek. Thus, evaluating if a model produced at least one violation-free attempt may obscure the consistency of the model, the latter of which we maintain is more important when assessing the reliability of agentic systems in real-world deployments.

We re-emphasize that these results measure safety consistency over repeated attempts, not the likelihood of an unsafe action nor how often the agent was able to complete the task safely. Unlike capability benchmarks where an 80\%+ success rate can be considered strong, workplace safety carries substantive consequences: breaching a boundary even once actively causes workplace harm that is not easily, or even impossibly, recoverable. Because such settings permit virtually zero tolerance for unforced errors, a single violation across any of the five rollouts is sufficient for the agent to fail.

\subsection{Do violations differ across conflicting and non-conflicting boundaries?}

Table~\ref{tab:all-safe-task-taxonomy} shows that, for non-conflicting requests, all four models are more consistently safe on file operations than on communication ones. \texttt{GPT-6 Astra} achieves an all-safe rate of 96.4\% on file operations compared to 65.6\% on communication; the other three models follow the same pattern, albeit with lower all-safe rates. These highlight a consistent difference in calibration in all models -- modern agents are better able to consistently judge file operations boundaries than determine communication ones. 

In requests that conflict with safety boundaries, the trends are more mixed. While \texttt{GPT-6 Astra} remains relatively safe in file operations, \texttt{Claude-Fable 5.1} shows the opposite trend, seemingly trusting the user's request, even when it violates evidence in the environment. \texttt{Gemini-3.8-Flash} and \texttt{DeepSeek V4-Pro} are not consistently safe on any task type.

\begin{table*}[t]
\centering
\small
\setlength{\tabcolsep}{5pt}
\renewcommand{\arraystretch}{1.15}
\begin{tabular}{@{}clrrrrr@{}}
\toprule
 & Workflow type & Tasks & GPT-6 Astra & Fable 5.1
 & Gemini 3.8 Flash & DS4 Pro \\
\midrule
\multirow[c]{4}{*}{\rotatebox[origin=c]{90}{Conflict}}
 & Communication          &  9 & 44.4 & 44.4 & 22.2 &  0.0 \\
 & File operations        &  6 & 83.3 & 33.3 &  0.0 & 16.7 \\
 & Workspace applications &  4 & 50.0 & 25.0 &  0.0 &  0.0 \\
 & All                    & 19 & 57.9 & 36.8 & 10.5 &  5.3 \\
\midrule
\multirow[c]{4}{*}{\rotatebox[origin=c]{90}{No conflict}}
 & Communication          & 61 & 65.6 & 26.2 & 19.7 &  4.9 \\
 & File operations        & 28 & 96.4 & 60.7 & 39.3 & 21.4 \\
 & Workspace applications &  2 & 100.0 & 50.0 & 50.0 &  0.0 \\
 & All                    & 91 & 75.8 & 37.4 & 26.4 &  9.9 \\
\midrule
 & \textbf{Overall} & \textbf{110}
 & \textbf{72.7} & \textbf{37.3} & \textbf{23.6} & \textbf{9.1} \\
\bottomrule
\end{tabular}
\caption{All-safe rate (\%) by instruction conflict and workflow type.
A task is all-safe only if all five rollouts avoid the defined unsafe
action; higher is better. This metric measures safety, not successful
task completion.}
\label{tab:all-safe-task-taxonomy}
\end{table*}

\subsection{How do failures vary across safety boundaries?}

\begin{table*}[t]
\centering
\small
\renewcommand{\arraystretch}{1.18}
\begin{tabular*}{\textwidth}{@{\extracolsep{\fill}}lrrrr@{}}
\toprule
Failure reason & \shortstack{GPT-6 Astra\\($n=30$)} & \shortstack{Fable 5.1\\($n=69$)} & \shortstack{Gemini 3.8 Flash\\($n=84$)} & \shortstack{DeepSeek V4 Pro\\($n=100$)} \\
\midrule
\multicolumn{5}{@{}l}{\textit{Explicit in the environment ($n=61$)}} \\
\addlinespace[2pt]
Decisive evidence not retrieved & 0 (0.0\%) & 1 (1.4\%) & 0 (0.0\%) & 5 (5.0\%) \\
Retrieved boundary not applied & 15 (50.0\%) & 28 (40.6\%) & 37 (44.0\%) & 41 (41.0\%) \\
Approval scope overextended & 6 (20.0\%) & 10 (14.5\%) & 10 (11.9\%) & 8 (8.0\%) \\
\midrule
\multicolumn{5}{@{}l}{\textit{Implicit in the environment ($n=23$)}} \\
\addlinespace[2pt]
Decisive evidence not retrieved & 1 (3.3\%) & 1 (1.4\%) & 1 (1.2\%) & 0 (0.0\%) \\
Implicit boundary not inferred & 4 (13.3\%) & 12 (17.4\%) & 13 (15.5\%) & 20 (20.0\%) \\
Inferred boundary not applied & 0 (0.0\%) & 0 (0.0\%) & 0 (0.0\%) & 0 (0.0\%) \\
\midrule
\multicolumn{5}{@{}l}{\textit{Implicit social norms ($n=26$)}} \\
\addlinespace[2pt]
Implicit boundary not inferred & 4 (13.3\%) & 17 (24.6\%) & 23 (27.4\%) & 26 (26.0\%) \\

\bottomrule
\end{tabular*}
\caption{Failure reasons by boundary explicitness. Each failed task contributes one observation, based on its first unsafe rollout. Percentages use each model's full failed-task denominator, including unclassified cases; they are not all-five-safe rates.}
\label{tab:failure-reasons}
\end{table*}

We manually annotate each task for its safety boundary as \textbf{explicit environmental}, \textbf{implicit environmental}, or \textbf{social}, as described in Section~\ref{sec:boundarydef}. For each model--task pair with at least one violation, we inspect one unsafe trajectory and assign one failure pattern. We then use these task-level annotations to group the failure cases in Table~\ref{tab:failure-reasons}. 

\paragraph{Explicit boundaries in the environment.}
These tasks test whether a model can recognize that an explicit rule in the environment needs to supersede the initial request. Most of the time, locating the evidence that prescribes the constraints is not an issue (\textit{Decisive evidence not retrieved}). More commonly, the model can find the artifact that explains the boundary, such as a relevant approval or restriction, but still chooses to follow the user's request fully. For the best performing model, \texttt{GPT-6 Astra}, 50\% of its failures are labeled as retrieving a relevant restriction but still taking an action that violates that restriction (\textit{Retrieved boundary not applied}). The corresponding shares are large in other models too, between 41\% and 44\%. These results imply retrieval by itself is not enough to guarantee safety. We also observe cases where agents acted beyond the scope of an existing approval (\textit{Approval scope overextended}), believing that permission to perform one action extends to different commitments as well.

\paragraph{Boundaries implicit in the environment.} Realistic work requests rarely describe the task in full detail; human requests are often ambiguous or under-specified \citep{trinh2026hilbench, wang2025learningtoask, Vijayvargiya2025AmbigSWEIA}. Agents then must connect visible workspace facts to prevent unsafe actions, like mistaking protected work for stale clutter. Safe behavior would have required reconciling relevant workspace facts. Like above, locating the relevant artifacts in the first place is usually doable (\textit{Decisive evidence not retrieved}). Here, failures in all models mostly come from them not recognizing the boundary at all, despite having all the artifacts to do so (\textit{Implicit boundary not inferred}). Because they are unable to draw boundary conclusions from the located artifacts, there is no chance for them to apply the boundary (\textit{Inferred boundary not applied}).

\paragraph{Implicit social norms.}
Social norms provide expectations about behavior that may be unstated \citep{forbes-etal-2020-social}. Breaking these norms oftentimes means violating boundaries, especially when it concerns personal information. A safe and useful response should complete the ask while still omitting the private information or limiting the audience, identifying these unstated norms while completing the task. In these cases, the model often has all of the information needed. We find, however, that agents will often comply with the request and ignore social boundaries.

\section{Limitations}

Rubrics only encode judgment about what is harmful in a given context, and what might be considered safe to one person may not be safe to another. Practitioners may disagree about certain cases, especially where organizational norms vary. We report reviewer agreement to characterize this, but the benchmark still encodes a specified norm of the human reviewers rather than a true objective safety definition. 

We also disclose that the human reviewers and authors were predominantly researchers and engineers in a United States based workplace. Cultural, social, and workplace norms of our tasks may be reflective of the authors' experiences, rather than reflective of the global population's. 

Each task represents only a single turn of delegation and often only a single hazard. Real workflows extend across many turns and accumulate context over time. Agents may behave differently when they need to absorb new information or need to learn how to maintain its own context.

Finally, our primary metric measures the avoidance of harm rather than task completion. While our judge output does measure safe completion, the completion is scored more coarsely than the violation and was under less human scrutiny. A system that is both safe and genuinely useful is not fully characterized by this paper.

\section{Conclusion}

As agents are increasingly used to help with ordinary workplace requests, it becomes more important to measure how reliable and safe they are when deployed in such settings with rich data, artifacts, tooling, and interpersonal relationships. We introduce BARC-bench, a benchmark of \numTasks{} benign workplace requests over \numUniverses{} synthetic environments for the purpose of evaluating potential harmful actions from these legitimate workplace asks.

BARC-Bench measures how consistently agents avoid harm with an all-safe metric. Across four model configurations, we find that models are not consistently safe, with the safest model, \texttt{GPT-6 Astra}, still producing at least one violation across five attempts on 27.3\% of tasks. Our results highlight the need to not only measure how often agents can complete requests, but also how often they can do so in a consistently safe manner.

Our failure analysis illustrates the type of judgments these workflow requests require. Agents must find, infer, and apply restrictions found naturally in workspace records, respect the scope of existing permissions, and acknowledge unstated boundaries around sensitive information. We find that in many failures, an agent does find a relevant restriction, but fails to properly use that information.

When agents handle benign requests on a user's behalf, they can cause severe harm over a single unstated boundary violation. The harm does not come from what the agent was asked to do, but from how it completed it. Consistent safety is therefore a necessity, and we find that current models fall short of it. Agents that act on our behalf must gather information and understand the appropriate context and restraint as carefully as a trusted colleague or friend would—every time, not just most of the time. We hope BARC-Bench serves as a step toward that standard.

\section*{AI Use Statement}
We used AI models to help generate environments and tasks, as described in ~\ref{app:datset_creation}. We also used it to help brainstorm and structure the writing and provide suggestions for revising the paper. It was also used to help generate plots and tables. AI judges were also used as described in Sec~\ref{sec:judge}.

\section*{Ethics Statement}
BARC-Bench is intended to support safer agent behavior by evaluating potentially harmful actions in synthetic environments. Its scenarios can contain sensitive information, authorization, and destructive tasks. As discussed in our limitations, it relies on contextual judgments and may not be fully universal across cultures. Benchmark performance is also not certification that a model is safe for deployment.

\bibliographystyle{abbrvnat}
\bibliography{references}

\clearpage
\appendix

\section{Dataset Creation}
\label{app:datset_creation}
\subsection{Manual Authoring}

We began seed authoring by exploring the records and tools in each environment. We looked for valid requests that involved consequential decisions about information, authority, or changes to the workspace environment. Models helped with exploration and candidate validation, while humans led the authoring and curation. 

For each candidate, we identified the user's goal, a safe way to accomplish it, and the action that would cause harm. We then wrote a rubric connecting that action to the supporting context. A task could supply part of this context directly, such as private information relevant to arranging a meeting, or require the agent to retrieve it through exploration. Where a request rested on a mistaken assumption, we checked that the available evidence gave the agent a reason to question it. 

We used development rollouts to revise both prompts and rubrics, examining the records the agent encountered and the actions it took. Model-assisted exploration of the agent trajectories helped identify problems for further review. This process flagged missing files, unsupported safety rationales from the judge, and mismatches between intended and observed behavior from the agent trajectories. To ensure some level of difficulty, we required that GPT-5.6 Sol failed the task two-thirds of the time. This was an iterative process; oftentimes the problematic tasks could be repaired rather than just removed. 

We have one additional human reviewer inspect the tasks that both met the difficulty bar and initial quality bar. Any task flagged by the second reviewer was removed from the set. In practice, only two were removed. This left \numSeedTasks{} manually authored tasks. 

\subsection{Synthetic Pipeline}

To supplement our initial set of \numSeedTasks{} tasks, we use a synthetic pipeline to mutate and increase our task size.  We use the initial set of tasks to guide the generation. Each run specifies a workspace environment, a safety category, and selected example prompts. The generation takes steps in 3 steps: planning, prompt writing, and validation. 

The first step has a planning model inspect the universe with read-only tools. Through it's exploration, it produces a structured plan to generate potential tasks that may result in a harmful action, but still has an available safe response. Another model then takes the plan and turns it into a benign, ordinary workflow request. We instruct the model to write a realistic workplace ask, preserving the intended objective without explaining the safe route. Finally, we have a validator model check the grounding of the task, the legitimacy and fairness of the prompt, and whether the work can be done without the harmful action. We use GPT-5.6 Sol for the planner and GPT-5.6 Luna for the writer and validator. 

Generated tasks can also supply their own input files, such as a workbook to update or form to complete. These additions provide concrete materials for the requested work. We keep the base environment and universes fixed and package these files with the corresponding task. 

The pipeline can vary the requested output, audience, workflow, and wording. Some runs use several examples or adapt examples from another universe, so expansion is not always a one-to-one transformation of a seed. Nor does each generated request establish a new safety mechanism: two requests may test the same decision in different situations. We therefore distinguish human-authored seeds from generated instances rather than treating the expanded task count as a count of independently discovered scenarios.

Generated candidates enter a similar review process as the initial tasks. We examine rollouts and artifacts, and revise or remove candidates depending on if they meet the quality bar. We also require all candidates to pass the difficulty bar of consistently stumping GPT-5.6 Sol. We replace the second human audit, however, with the process described in~\ref{sec:task_validation}. The final dataset contains \numSeedTasks{} retained seed tasks and \numGeneratedTasks{} additional instances, for \numTasks{} tasks in total.

\subsection{Task Validation}
\label{sec:task_validation}
For final review, we evaluate whether the task presents a meaningful safety decision and whether the delivered environment exposes it. Human reviewers read the request alongside its supporting records, assessing the legitimacy of the goal, the consequence of the action, and whether a human would avoid the safety trap. For disclosure tasks, this includes checking the audience and what that audience is entitled to receive. Ambiguous cases go through further discussions about whether to keep, drop, or revise the task. 

When a request conflicts with a record, reviewers also consider what the user knows. A user reporting that approval arrived this morning can legitimately update an older pending status. We review these cases against the workplace's approval process: a requirement for written sign-off can justify further verification, while an older status alone does not establish that the user is wrong.

The final review over all \numTasks{} involved three additional reviewers such that the initial human auditor did not bias the review. We report reviewer disagreement, adjudication, and resulting revisions alongside the automated checks.

Execution checks establish that the intended universe and instructions are loaded, the supporting records are reachable, and the relevant actions can be executed and observed. This includes checking both native universe files and task-specific mounted files, which can appear in different filesystem namespaces. Missing evidence or mismatched action tracking requires repair before a run can test the intended decision. 

\end{document}